\documentclass[conference]{IEEEtran}
\IEEEoverridecommandlockouts
\usepackage{cite}
\usepackage{amsmath,amssymb,amsfonts}
\usepackage{algorithmic}
\usepackage{graphicx}
\usepackage{textcomp}
\usepackage{xcolor}
\usepackage[T1]{fontenc}

\usepackage[pagebackref=true,breaklinks=true,letterpaper=true,colorlinks,bookmarks=false]{hyperref}

\def\BibTeX{{\rm B\kern-.05em{\sc i\kern-.025em b}\kern-.08em
    T\kern-.1667em\lower.7ex\hbox{E}\kern-.125emX}}
    \IEEEoverridecommandlockouts
\IEEEpubid{\makebox[\columnwidth]{978-1-7281-9023-5/21/\$31.00~\copyright{} 2021 IEEE \hfill}
\hspace{\columnsep}\makebox[\columnwidth]{ }}
\begin{document}

\title{Multiple Object Trackers in OpenCV: A Benchmark
}

\author{\IEEEauthorblockN{Nađa Dardagan, Adnan Brđanin}
\IEEEauthorblockA{\textit{Department of Informatics} \\
\textit{Alpen-Adria-Universität Klagenfurt}\\
Klagenfurt am Wörthersee, Austria \\
\{n1dardagan, adnanbr\}@edu.aau.at}
\and
\IEEEauthorblockN{Džemil Džigal, Amila Akagic}
\IEEEauthorblockA{\textit{Department of Computer Science and Informatics} \\
\textit{Faculty of Electrical Engineering, University of Sarajevo}\\
Sarajevo, Bosnia and Herzegovina \\
\{ddzigal1, aakagic\}@etf.unsa.ba}
}

\maketitle

\begin{abstract}
Object tracking is one of the most important and fundamental disciplines of Computer Vision. Many Computer Vision applications require specific object tracking capabilities, including autonomous and smart vehicles, video surveillance, medical treatments, and many others. The OpenCV as one of the most popular libraries for Computer Vision includes several hundred Computer Vision algorithms. Object tracking tasks in the library can be roughly clustered in single and multiple object trackers. The library is widely used for real-time applications, but there are a lot of unanswered questions such as when to use a specific tracker, how to evaluate its performance, and for what kind of objects will the tracker yield the best results? In this paper, we evaluate 7 trackers implemented in OpenCV against the MOT20 dataset. The results are shown based on Multiple Object Tracking Accuracy (MOTA) and Multiple Object Tracking Precision (MOTP) metrics. 
\end{abstract}

\begin{IEEEkeywords}
Computer Vision, Object Tracking, Multiple Object Tracking, OpenCV, Benchmark
\end{IEEEkeywords}

\section{Introduction}
Object tracking is one of the fundamental Computer Vision tasks. Despite the progress made in the field over the last decade, object tracking is still considered a very challenging task due to numerous factors that can affect a tracker's overall performance, such as background clutter, occlusion, illumination variation, and others~\cite{wu2013online}. Creating a tracker that could perform well enough in various situations is still considered an unsolved problem. Thus, it is important to understand how to choose the right tracker for the problem at hand.

%Therefore, it is crucial to evaluate the performance of state-of-the-art trackers to demonstrate their strength and weakness and help identify future research directions in this field for designing more robust algorithms

The complexity and performance of object tracking algorithms often depend on how many objects are tracked in a video sequence. Object tracking tasks can be classified based on how many objects are being tracked in a sequence into a) single object and b) multiple objects trackers. A Single Object Tracking (SOT) algorithm tracks only a single object in a video sequence, and it is successful if it tracks an object even if the environment consists of multiple objects. In our previous work~\cite{brdjanin2020single}, we have benchmarked SOT trackers in the OpenCV and provided some insights on how to choose a tracker for a specific type of task.

In this paper, we review Multiple Object Tracking (MOT) algorithms by analyzing frame by frame of multiple video sequences to identify all objects on the scene and how their position changes over time. The output from every frame is a bounding box for every object of interest which uniquely defines its position in the current frame. The object ID is added to the returned bounding box to identify the object of interest. The main difference to Single Object Tracking is that in SOT the appearance of the target object is known in advance, while in MOT it is necessary to detect the target. The difficulty of multiple object tracking is that targets can appear and disappear from the scene, which introduces the problem of figuring out whether the reappeared object on the scene was already seen or a completely new object. This is because objects can have very similar appearances. These difficulties are also a reason why using SOT algorithms to track multiple objects is not a good idea and it is not uncommon for objects to switch IDs.

The object trajectory can be reconstructed by analyzing the returned bounding boxes with the object ID. There are numerous applications for such systems, from video surveillance, smart vehicles to medical treatments. This is why MOT attracts a lot of attention in the past years and as a result, many object tracking algorithms were proposed. In recent years, object tracking community embraced data-driven learning methods, following the rise of deep learning in computer vision~\cite{muller2018trackingnet}, \cite{ciaparrone2020deep} \cite{xu2019deep}. In this paper, we assess the quality of multiple object tracking algorithms implemented in the OpenCV library. The library was chosen because of its simplicity, versatility, efficiency, and the fact that it is one of the biggest developing open-source computer vision libraries. Several approaches and methods for MOT make it hard to overview their performance and distinguish what are suitable usages. That is why we decided to benchmark seven algorithms available in OpenCV.

The structure of the paper is the following: In Section~\ref{rw} we provide a short overview of datasets developed for MOT. In Section~\ref{overview}, the concept of Multiple Object tracking is provided. In Section~\ref{evaluation}, we explain the evaluation metrics and methods we used to measure the performance of algorithms. Section~\ref{dataset} contains details about the chosen dataset. Finally, in Section~\ref{results} we summarize the results of benchmarking. Section~\ref{conclusion} provides a conclusion and some recommendations about how to choose an appropriate tracker.  

\section{Related Work}\label{rw}
Many different object tracking datasets have been proposed in recent literature, due to the importance and interest of the topic. However, there is still no standardization on how to evaluate a tracker so different metrics are used. The metrics combined can give clearer insight on when to use which tracker and how well each of the algorithms performs. From a number of object tracking benchmarks, such as~\cite{wu2013online}~\cite{milan2016mot16}~\cite{fan2019lasot}~\cite{muller2018trackingnet},  classification of datasets can be made based on the annotation density:

\begin{enumerate}
    \item \textbf{Dense} - every frame is manually labeled with careful inspection;
    \item \textbf{Other} - sequences are sparsely and/or semi-automatically annotated (i.e. each frame might not be annotated). 
\end{enumerate}

Some notable object tracking evaluation datasets include KITTI~\cite{geiger2012we}, VOT~\cite{kristan2015visual}~\cite{kristan2017visual}, Need for Speed~\cite{kiani2017need}, Trackingnet~\cite{muller2018trackingnet}, Cdtb~\cite{lukezic2019cdtb}, Lasot~\cite{fan2019lasot}, Got-10k~\cite{huang2019got}, Jrmot~\cite{shenoi2020jrmot}, which are sorted by the year of publication.

To the best of our knowledge, a full benchmark of the multiple object trackers implemented in OpenCV is still not performed. This paper gives an insight into how well they perform and what's the ideal use-case for each of them.

Our previous paper~\cite{brdjanin2020single} contains details about the OpenCV library and benchmark results of single object tracking algorithms.

\section{Overview of Multiple Object Tracking}
\label{overview}
In this section, we focus on multiple object tracking algorithms in the OpenCV library. The library contains seven algorithms for MOT in the Multitracker class, accessible through the OpenCV API. The information about the algorithms, including references to research papers, may be found in table \ref{table:1}. Multi tracker class in OpenCV is implemented as a vector of single object trackers. This implementation makes measuring the speed of an algorithm more complicated than in SOT algorithms. The speed of tracking objects or the FPS (Frames per Second) of a single video is not possible to calculate, since every tracker has its own FPS and it is difficult to assess how much a certain tracker contributes to the overall speed of a video. 

The next section explains the evaluation metrics and methods implemented for measuring the performance of the trackers. 

\begin{table*}
\centering
\caption{OpenCV multiple object trackers sorted by the year of their publication. Google Scholar Citations are accessed on January 31st 2021.}
\label{table:1}
\begin{tabular}{|l|l|l|l|} 
 \hline
No & Tracker Full Name & Publication Title and & Publication Year \\
 &  (Abbreviation) & Reference  & (Google Scholar \\ 
  &   &   & Citations) \\ \hline
1. & \textbf{Boosting} &Real-time tracking via on-line boosting~\cite{grabner2006real} & 2006 (1555) \\ \hline 
2. & Multiple Instance Learning (\textbf{MIL}) & Visual tracking with online multiple instance learning~\cite{babenko2009visual}& 2009 (2259)\\ \hline
3. & \textbf{MedianFlow} &Forward-backward error: Automatic detection of tracking failures~\cite{kalal2010forward}&  2010 (925)\\ \hline
4. & Minimum Output Sum of Squared Error (\textbf{MOSSE}) & Visual object tracking using adaptive correlation filters~\cite{bolme2010visual} & 2010 (2319)\\ \hline
5. & Tracking Learning Detection (\textbf{TLD}) & Tracking-learning-detection~\cite{kalal2011tracking}& 2011 (3667) \\ \hline
6. &  Kernelized Correlation Filter (\textbf{KCF}) & High-speed tracking with kernelized correlation filters~\cite{henriques2014high} & 2014 (3895)\\ \hline  
7. &  \textbf{CSRT} (Channel and Spatial Reliability Tracker) & Discriminative Correlation Filter with Channel and Spatial Reliability~\cite{lukezic2017discriminative} & 2017 (674)\\ \hline 
\end{tabular}
\end{table*}

\section{Evaluation}
\label{evaluation}
Choosing the right evaluation metric is of high importance to ensure that no important information about an individual algorithm has been lost. Based on~\cite{dendorfer2020mot20}, two evaluation metrics are chosen: Multiple Object Tracking Accuracy (MOTA) and Multiple Object Tracking Precision (MOTP). The MOTA shows if the tracker found the right objects in a frame, while MOTP represents how successful was object tracking. The average accuracy and precision are obtained by the execution of every tracking algorithm on test data, where fifty random objects were selected to be tracked from the ground truth positions of the objects in the first frame of the corresponding videos. To evaluate how different trackers behave under different conditions One Pass Evaluation (OPE) method was used. 

\subsection{Multiple Object Tracking Accuracy}
The MOTA is based on intersection over union as a distance measure between the ground truth bounding box and a tracker output. Intersection over Union is calculated for each object in every frame and it takes values from [0, 1]. The average accuracy of a tracker is the sum of average accuracies of all tracked objects in a test sequence divided by $50$ (in this case, $50$ is the number of tracked objects). Tracker has a good accuracy if the value is above a certain threshold $t$, where $t = 0.5$.

\subsection{Multiple Object Tracking Precision}
The MOTP is based on the Euclidean distance between centers of the ground truth bounding box and the trackers bounding box of every object in every frame of the test video. Similar to the average MOTA, the average MOTP is the sum of average precisions of all tracked objects in a test sequence divided by $50$. 
The threshold of the precision value is set to $t = 50px$, meaning that every tracker with an average distance less than $t$ has good precision. 

\subsection{One Pass Evaluation method} 
The OPE is a method for evaluating the robustness of a tracker to different sizes of initial bounding boxes. Each tracker is initialized for every object with a bounding box from a ground truth in a starting frame. The algorithm tracks an object until it leaves the frame and no reinitialization is done after the failure of the tracker. These cases are detected with a value of intersection over union being zero. OPE is the simplest robustness evaluation method since it does not show the algorithm's vulnerability to different types of initialization or state of the object in the initial frame.

In this paper, we evaluate the following Multi-Object Tracking algorithms from the OpenCV library: Boosting, MIL, MedianFlow, MOSSE, TLD, KCF, and CSRT.  

\begin{figure*}
\centering
\includegraphics[height=3cm, width=5.5cm]{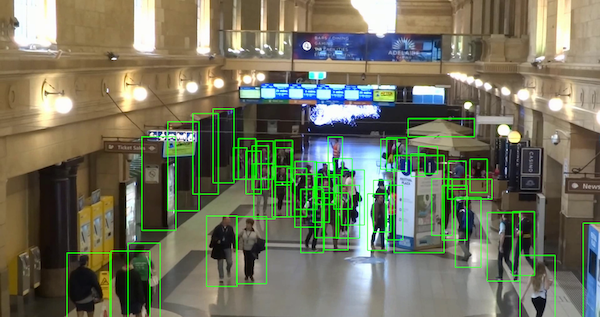} 
\includegraphics[height=3cm, width=5.5cm]{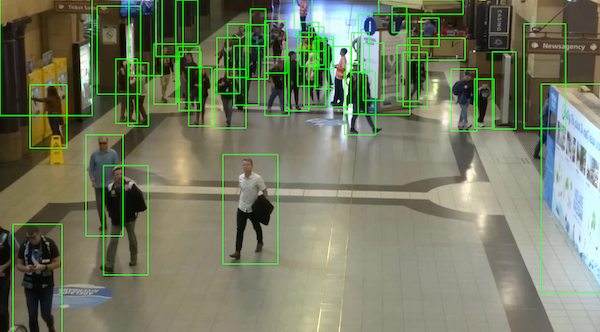} \\  
\includegraphics[height=3cm, width=5.5cm]{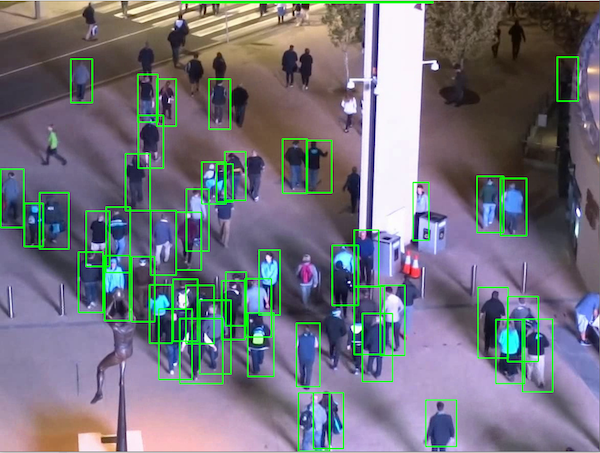} 
\includegraphics[height=3cm, width=5.5cm]{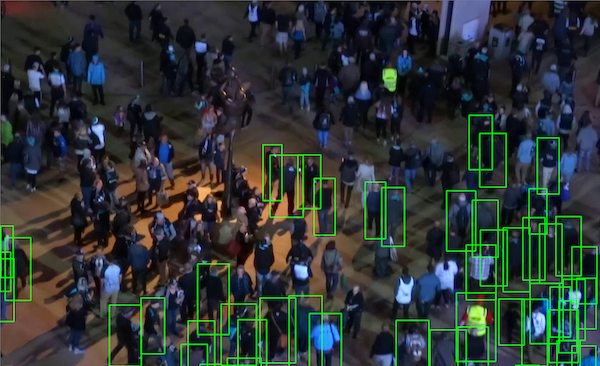}
\caption{Example scenes from the dataset while tracking 50 objects with CSRT algorithm. Best viewed in color.}
\label{fig:datasetExample}
\end{figure*}

\section{Dataset}
\label{dataset}
This section reviews the dataset used for the evaluation of multiple object trackers in OpenCV. We use \textbf{MOT20}\footnote{For more information about the dataset visit \url{https://motchallenge.net/data/MOT20/}} dataset. Multiple Object Tracking Challenge was first introduced in 2014 and serves as a competition for evaluating newly developed object tracking algorithms. MOTChallenge datasets have improved vastly over the last couple of years. Because of the previously mentioned multiple object tracking challenges, this dataset was developed to be fully annotated along with evaluation metrics to assess the quality of the algorithm. MOT20 dataset is one of the latest releases from MOTChallenge. The dataset is intended to be used for dense scene tracking, namely pedestrian tracking. It consists of eight videos, split into four test videos and four train videos. MOT20 dataset was chosen because it is precisely annotated, which means it strictly follows a well-defined annotation protocol. The dataset is challenging as the number of objects to track found in one frame can go up to $246$~\cite{dendorfer2020mot20}. All tracking logs are shown as image coordinates $(x_i, y_i, w_i, h_i)$ where $x_i$ and $y_i$ represent the coordinates of the upper left corner of the object bounding box and $w_i$ and $h_i$ represent the width and height of the bounding box respectively, where index $i$ represents $i$-th bounding box in the video. Taking all the dataset properties into account, we decided to use this dataset because of its complexity and annotation precision. The complexity of the algorithm evaluation on this dataset is reflected in the fact that the scenes are dense, with pedestrians being occluded at some point in time, coming in and out of the scene. 

All objects can be classified in three categories: 
\begin{enumerate}
    \item Moving pedestrians;
    \item People that are not in the upright position, non-moving pedestrians or artificial representations of humans;
    \item Vehicles and occluders.
\end{enumerate}

In our $50$ objects threshold for tracking, we encountered $4$ categories of objects: 
\begin{enumerate}
    \item Pedestrian (belongs to moving pedestrians category);
    \item Static Person (belongs to non-moving pedestrians category);
    \item Occluder full (belongs to vehicles and occluders category);
    \item Crowd (belongs to vehicles and occluders category).
\end{enumerate}

It is important to emphasize that we used the four training videos for our evaluation since it contains precisely annotated ground truth bounding boxes file, which we compared with the output of the OpenCV tracker. We did not use the detection file provided in the dataset, since this benchmark does not require it. The sample scenes from the dataset can be seen in Fig.~\ref{fig:datasetExample}. 

\begin{table*}
\centering
\caption{Dataset overview}
\label{table:dataset_overview}
\begin{tabular}{|l|l|l|l|l|l|}
    \hline
    Video Name & FPS & Duration     & Resolution  & Num. of boxes & Description                                 \\ \hline
    MOT20-01   & 25  & 429 (00:17)  & 1920 x 1080 & 26647         & Crowded indoor train station                \\ \hline
    MOT20-02   & 25  & 2782 (01:51) & 1920 x 1080 & 202215        & Crowded indoor train station                \\ \hline
    MOT20-03   & 25  & 2405 (01:36) & 1173 x 880  & 356728        & People leaving entrance of stadium at night \\ \hline
    MOT20-05   & 25  & 3315 (02:13) & 1654 x 1080 & 751330        & Crowded city square at night                \\ \hline
\end{tabular}
\end{table*}

\begin{figure}
    \centering
    %trim=left botm right top
    %\includegraphics[trim=0cm 4cm 0cm 4cm, width=0.5\textwidth, clip]{Figures/MOT20-01IoU.pdf}
    %\includegraphics[trim=0cm 9cm 0cm 7cm, width=0.5\textwidth, clip]{Figures/MOT20-01CD.pdf}
    \includegraphics[trim=0cm 4cm 0cm 0cm, width=0.5\textwidth]{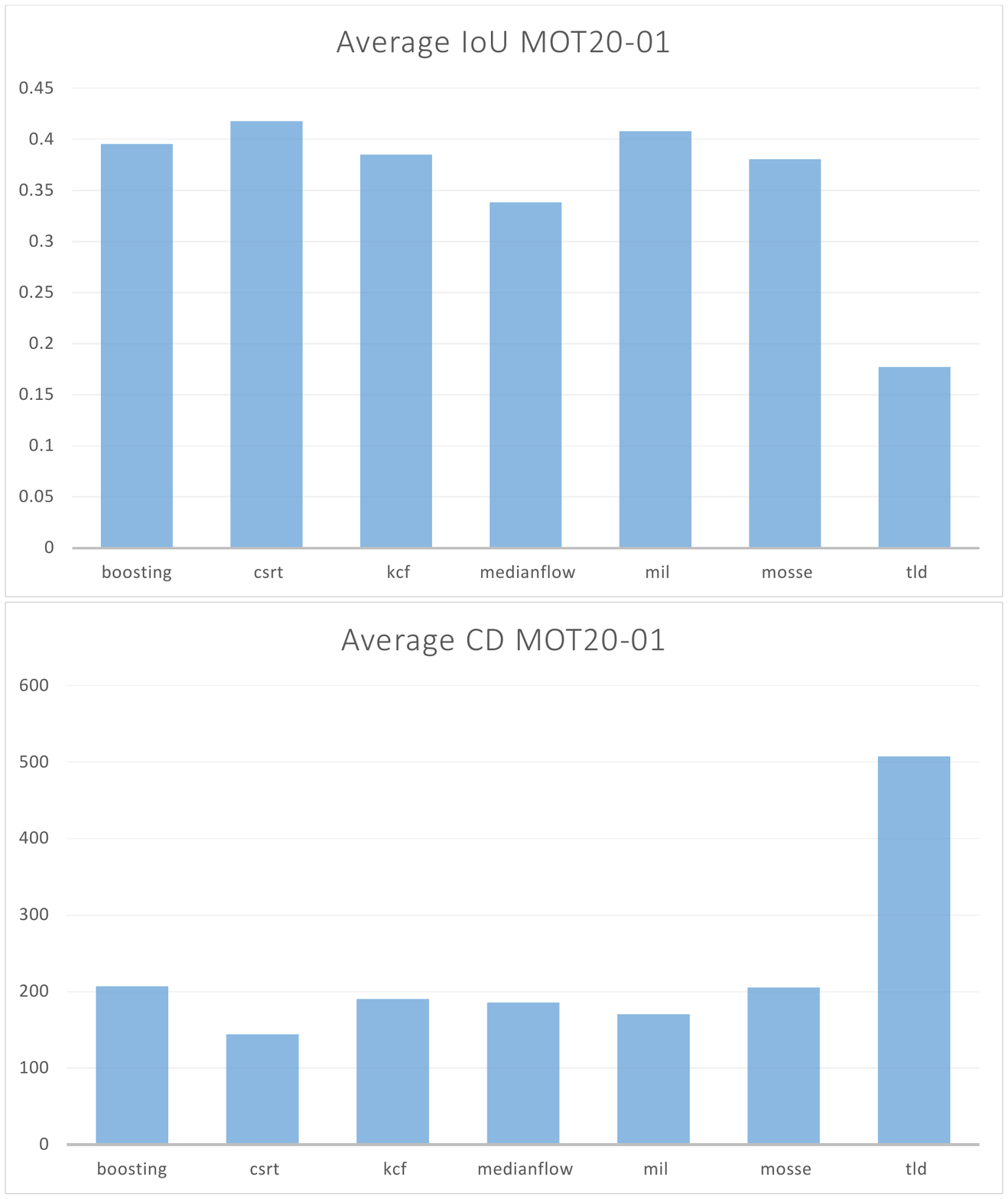}

    \caption{IoU and CD results for MOT20-01.}
    \label{fig:resultsPlot1}
\end{figure}
\begin{figure}
    \centering
    \includegraphics[trim=0cm 4cm 0cm 0cm, width=0.5\textwidth]{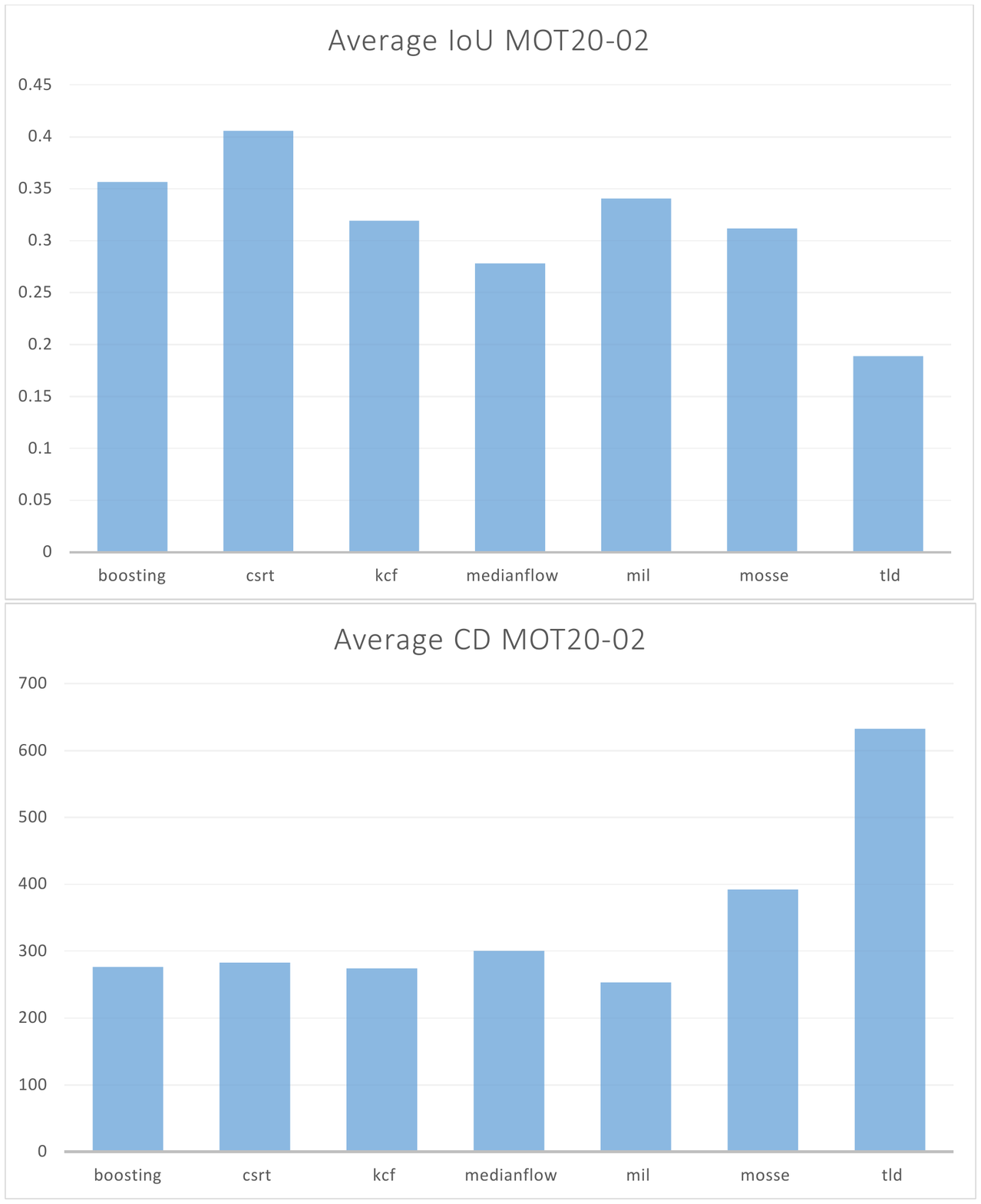}
    %trim=left botm right top
    %\includegraphics[trim=0cm 9cm 0cm 7cm, width=0.5\textwidth, clip]{Figures/MOT20-02IoU.pdf}
    %\includegraphics[trim=0cm 9cm 0cm 7cm, width=0.5\textwidth, clip]{Figures/MOT20-02CD.pdf}
    \caption{IoU and CD results for MOT20-02.}
    \label{fig:resultsPlot2}
\end{figure}

\begin{figure}
    \centering
    %trim=left botm right top
     \includegraphics[trim=0cm 4cm 0cm 0cm, width=0.5\textwidth]{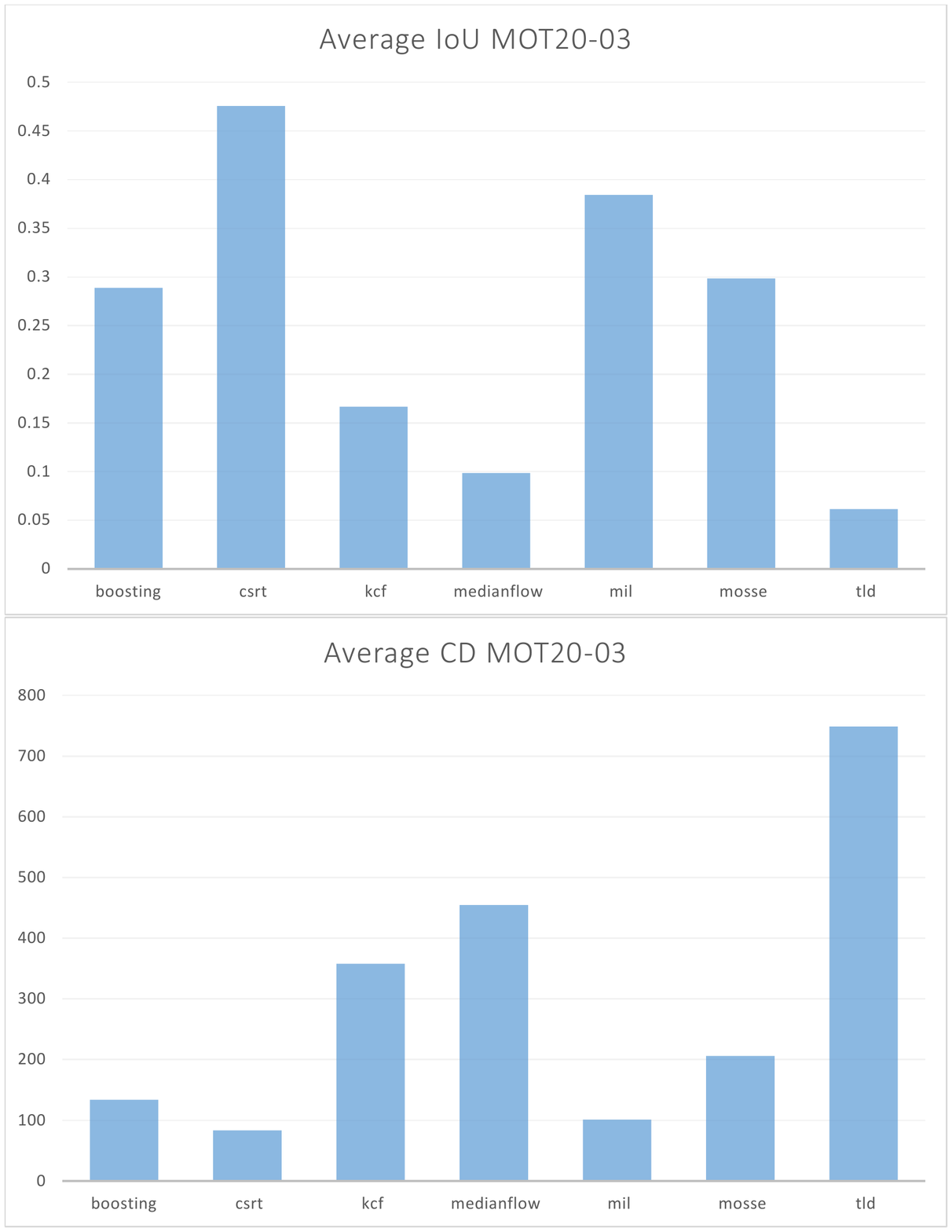}
    \caption{IoU and CD results for MOT20-03.}
    \label{fig:resultsPlot3}
\end{figure}
\begin{figure}
    \centering
    %trim=left botm right top
     \includegraphics[trim=0cm 4cm 0cm 0cm, width=0.5\textwidth]{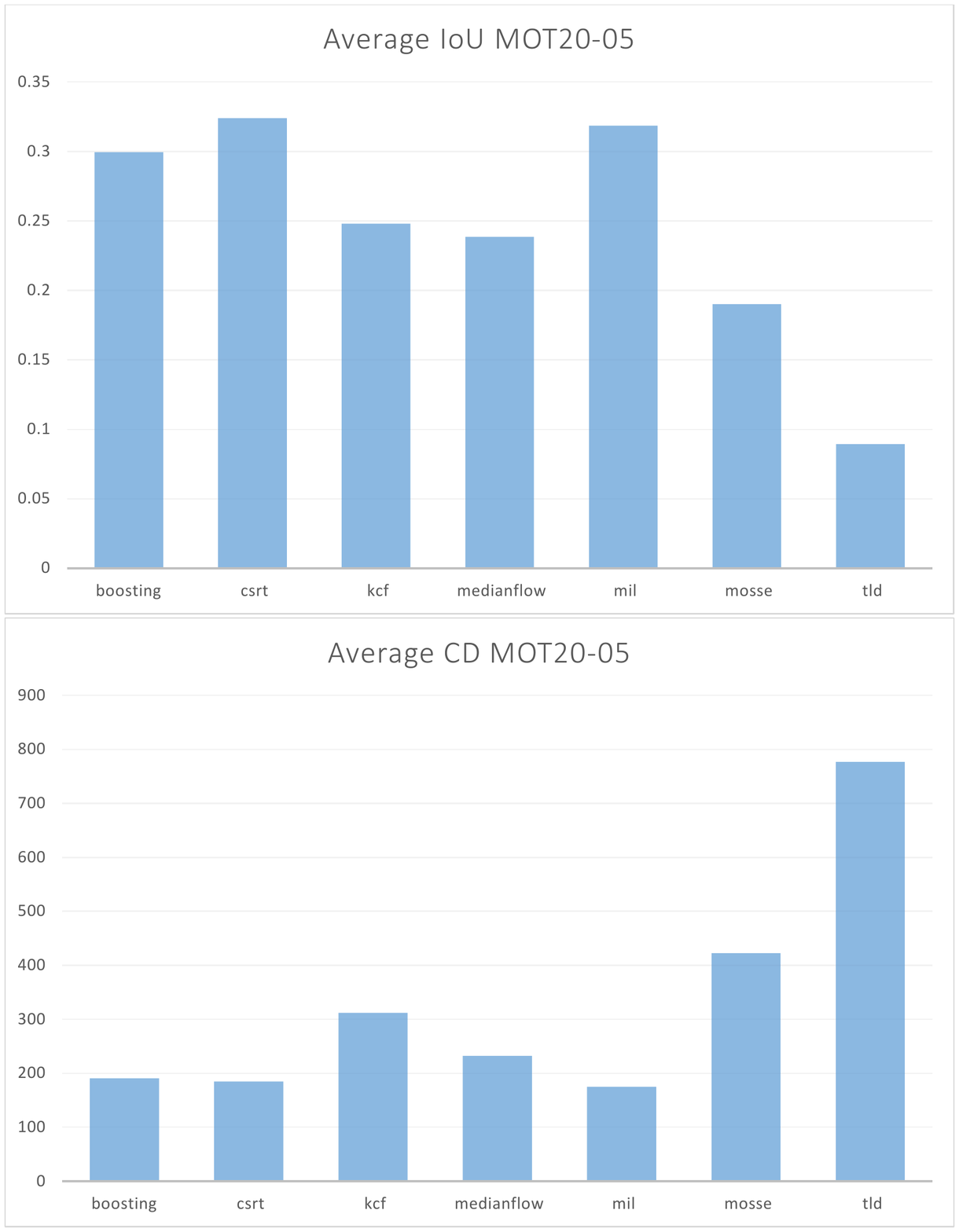}
    \caption{IoU and CD results for MOT20-05.}
    \label{fig:resultsPlot4}
\end{figure}

\section{Results}
\label{results}
All tracker performances were evaluated on a Windows-based machine with i7-8565U CPU @ $1.80$ GHz with four cores. The evaluation tests were successfully executed for all trackers. It is important to note that the speed of the tracking algorithms was not calculated precisely, because of the OpenCV implementation of multiple trackers. As mentioned previously, OpenCV implements a multiple object tracker as an array of single object trackers which makes it difficult to calculate the exact frames per second rate. When two objects of a different type, one static and the other one dynamically changing its position, are tracked with one multiple object tracking algorithm, each single object tracker from the array of trackers might have a significantly larger/smaller FPS rate. Single tracker tracking a static object requires less computational resources which implies a higher FPS rate. On the other hand, a single tracker tracking an object that changes its position quickly over time requires more computational power and results in a smaller FPS rate. With this said, taking the average of all FPS rates as an average speed of the multiple object tracker is not correct and might give falsified results. An example would be tracking 40 static and 10 dynamic objects which would yield a higher FPS rate than it actually is. With this in mind, we do not calculate the FPS precisely, but vaguely estimated if a multiple object tracking algorithm can be used in a real-time application. The estimate for the speed is as follows: none of the algorithms have the potential to be used in real-time scenarios except for MOSSE and MedianFlow. These trackers have successfully tracked all objects on the scene in real-time on the machine with the configuration stated previously, while other algorithms struggled when the number of objects in the scene was even less than $10$.

It is difficult to pinpoint one tracker that performs the best, so the evaluation will be done from two different perspectives: accuracy and precision, listing the trackers in the order from the best to the worst. 

%The first aspect of tracker evaluation is the duration of tracking. All evaluated OpenCV trackers, except for MOSSE and Medianflow, show no real-time application potential where the number of tracked objects is greater than $10$. However, MOSSE and Medianflow have proved to be speed-efficient even when it comes to $100+$ objects in one frame of the scene. 

The first aspect is the multiple object tracking accuracies (MOTA). This metric is based on Intersection Over Union, presented in the Evaluation Section. We considered that tracking is successful whenever the intersection over union ratio is not equal to 0. As soon as the IoU rate drops to $0$, the object is lost and none of the trackers successfully recovered to track the same object again. Average results are shown in Table \ref{table:iou}. CSRT has the best accuracy rate, followed by MIL, Boosting, MOSSE, KCF, Medianflow, and TLD. 

The second aspect of interest is the multiple objects tracking precision (MOTP). This metric is based on Center Distance measured in pixels. Average results are shown in Table \ref{table:cd}. CSRT has the best precision rate as well, followed closely by MIL and Boosting. KCF, Medianflow, MOSSE, and TLD have shown worse results, in the order in which they are stated. 

To conclude, if the focus is on accuracy and precision, CSRT is the best algorithm to use but the compromise is being made on the speed of object tracking. If the speed of tracking is the main concern, the recommendation is to use the MOSSE algorithm, keeping in mind the compromise made on precision and accuracy. If one is looking for an algorithm that balances both aspects, i.e. faster than CSRT and more precise than MOSSE, Boosting or MIL would be recommended. TLD performed the worst overall, considering all three factors, and therefore is not a recommendation to use for multiple object tracking\footnote{The code used for evaluation and the results we retrieved can be found on the GitHub repository: \url{https://github.com/adnanb97/OpenCV-Multiple-Object-Tracking}.}. 

\begin{table*}
\centering
\caption{Average Accuracy Rate for each tracker on all 4 train videos.}
\label{table:iou}
\begin{tabular}{|l|l|l|l|l|l|l|l|}
\hline
Video/Tracker & Boosting & CSRT   & KCF    & Medianflow & MIL    & MOSSE  & TLD    \\ \hline
MOT20-01      & 0.3955   & 0.4179 & 0.3852 & 0.3382     & 0.4079 & 0.3803 & 0.1771 \\ \hline
MOT20-02      & 0.3564   & 0.4057 & 0.3193 & 0.2783     & 0.3406 & 0.3118 & 0.1888 \\ \hline
MOT20-03      & 0.2888   & 0.4755 & 0.1667 & 0.0984     & 0.3843 & 0.2983 & 0.0614 \\ \hline
MOT20-05      & 0.2995   & 0.3240 & 0.2480 & 0.2385     & 0.3186 & 0.1901 & 0.0895 \\ \hline
\end{tabular}
\end{table*}

\begin{table*}
\centering
\caption{Average Precision Rate for each tracker on all 4 train videos.}
\label{table:cd}
\begin{tabular}{|l|l|l|l|l|l|l|l|}
\hline
Video/Tracker & Boosting & CSRT   & KCF    & Medianflow & MIL    & MOSSE  & TLD    \\ \hline
MOT20-01      & 206.75   & 144.06 & 190.41 & 185.55     & 170.61 & 205.24 & 507.56 \\ \hline
MOT20-02      & 276.45   & 282.84 & 274.19 & 300.43     & 253.08 & 392.33 & 632.71 \\ \hline
MOT20-03      & 134.03   & 83.61  & 358.19 & 454.53     & 101.43 & 205.93 & 748.42 \\ \hline
MOT20-05      & 190.81   & 184.65 & 312.03 & 232.36     & 175.12 & 422.48 & 776.68 \\ \hline
\end{tabular}
\end{table*}

\section{Conclusion}
\label{conclusion}
In this paper, we evaluate multiple object trackers available in the OpenCV library. We evaluate the algorithms based on their duration, accuracy, and precision, taking the threshold of $50$ objects tracked at every frame of the four videos in the MOT20 dataset. The evaluation is based on the class of the object, the label of the object, and conducted based on the well-annotated ground truth bounding boxes provided in the dataset. MOSSE and Medianflow have the potential to be used in real-time scenarios, where the number of tracked objects on a scene can exceed $100$, however, the tradeoff is in the precision of tracking. Other algorithms do not have the potential for real-time applications if the number of tracked objects on a scene exceeds $10$. With regards to accuracy and precision, the CSRT tracker performed the best followed by MIL and Boosting, which execute at a faster rate compared to CSRT, but with significantly lower accuracy and success rates. %We can conclude that these results align with our previous work~\cite{brdjanin2020single}.

\bibliographystyle{IEEEtran}
\bibliography{refs}

\end{document}